\documentclass{article}

\usepackage{arxiv}

\usepackage[utf8]{inputenc} 
\usepackage[T1]{fontenc}    
\usepackage{hyperref}       
\usepackage{url}            
\usepackage{booktabs}       
\usepackage{amsfonts}       
\usepackage{nicefrac}       
\usepackage{microtype}      
\usepackage{lipsum}		

\usepackage[table]{xcolor}
\usepackage{graphicx}
\usepackage{subfig} 
\usepackage{svg}
\usepackage{amsmath}
\usepackage{trimclip}
\clearpage{\thispagestyle{empty}\cleardoublepage}
\usepackage{adjustbox}
\usepackage{longtable}
\usepackage{float}
\usepackage{flafter}
\usepackage{stfloats}
\usepackage{listings}
\usepackage{color}
\usepackage{enumitem}

\title{Towards Using Machine Learning to Generatively Simulate EV Charging in Urban Areas}

\date{} 					

\author{%
  Marek Miltner\thanks{Alternative email: marek.miltner@cvut.cz} \\
  FEE CTU in Prague\\
  Prague, Czech Republic;\\
  CEE, Stanford University\\
    Stanford, USA\\
  \texttt{marek.miltner@stanford.edu} \\
  \And
  Jakub Zíka \\
  FEE CTU in Prague\\
  Prague, Czech Republic\\
  \texttt{zikajak3@fel.cvut.cz} \\
    \And
  Daniel Vašata \\
  FIT CTU in Prague \\
  Prague, Czech Republic\\
  \texttt{daniel.vasata@fit.cvut.cz} \\
  \And
  Artem Bryksa \\
  FEE CTU in Prague\\
  Prague, Czech Republic\\
  \texttt{bryksart@fel.cvut.cz} \\
  \And
  Magda Friedjungová \\
  FIT CTU in Prague\\
  Prague, Czech Republic\\
  \texttt{magda.friedjungova@fit.cvut.cz} \\
  \And
  Ondřej Štogl \\
  FEE CTU in Prague\\
  Prague, Czech Republic\\
  \texttt{stoglond@fel.cvut.cz} \\
  \And
  Ram Rajagopal \\
  CEE, Stanford University\\
  Stanford, USA\\
  \texttt{ramr@stanford.edu} \\
  \And
  Oldřich Starý \\
  FEE CTU in Prague\\
  Prague, Czech Republic\\
  \texttt{staryo@fel.cvut.cz} \\
}

\date{}

\renewcommand{\undertitle}{Accepted to Tackling Climate Change with Machine Learning: workshop at NeurIPS 2024.}

\begin{document}
\maketitle


\begin{abstract}
This study addresses the challenge of predicting electric vehicle (EV) charging profiles in urban locations with limited data. Utilizing a neural network architecture, we aim to uncover latent charging profiles influenced by spatio-temporal factors. Our model focuses on peak power demand and daily load shapes, providing insights into charging behavior. Our results indicate significant impacts from the type of Basic Administrative Units on predicted load curves, which contributes to the understanding and optimization of EV charging infrastructure in urban settings and allows Distribution System Operators (DSO) to more efficiently plan EV charging infrastructure expansion.
\end{abstract}


\section{Introduction and motivation}

One of the major pushes to fight climate change is the decarbonization of energy and mobility, which are closely related and together significantly contribute to global carbon emissions\cite{kais2016econometric}. Within this intersection, the electrification of mobility via large-scale deployment of electric vehicles (EVs) is one of the potential tools to decarbonize mobility in the coming decades\cite{ahmadi2019environmental, zhang2021long}. 

This push, however, requires significant investment in power infrastructure, mainly on the distribution level operated by distribution system operators (DSOs)\cite{dias2015multi, de2017joint}. This is due to the need to expand the availability of not only private but also public charging infrastructure, which can help better distribute loads across space and time\cite{gnanavendan2024challenges, un2017comprehensive}.

Since power engineering infrastructure expansion is an effort requiring significant time and financial resources, a critical challenge in this area is how to optimize grid expansion for efficiency to cover anticipated EV charging demand in various areas while not overloading the current network and not overspending on areas where demand is not as high\cite{powell2022charging,zhang_factors_2018}. This is especially difficult for DSOs since there has been a general lack of studies demonstrating analysis of real-world EV charging data in different geographies, mainly due to the data being vendor-locked and treated as confidential\cite{bhattarai2019big, potdar2018big, powell2022charging}. 

In this study, we aim to fill this critical gap by collaborating with PREdistribuce, the DSO in Prague, and the largest operator of public EV chargers. We propose a method to generatively create anticipated EV charging load curves based on location characteristics even in places where no chargers are present, in order to allow for DSOs and the wider community to better understand and simulate public EV charging behavior.


\section{Generative modeling of public charging demand}

In order to solve the challenge of predicting charging profiles for selected locations with limited information, we first analyzed the information about the landscape of public chargers in Prague. Our study is based on data sourced from public EV chargers operated by PREdistribuce, the DSO in the Prague area. 

We have gotten access to full logs of all charging sessions, including precise start and end times, power consumption, type and power output of each charger, and their location. State of charge or identification of individual vehicles was not available to us. We have paired this information with other geographic and demographic data to complement our understanding of the individual charger locations, based around the basic administrative units the chargers are located in. The public EV charging landscape is further explored in Appendix \ref{sec:app-a}. 

\subsection{Model inputs}
For the generative predictions of load curves, we have assessed several factors that might come into play to affect charging loads\cite{powell2022charging}. In connection to the basic administrative units, we were able to include several attributes that, in our opinion, are able to increase the performance of our model, including the area character and the amounts of local and long-haul commuters. Firstly, to reflect the character of the surrounding area, we include the basic administrative unit type, which includes 12 distinct categories ranging from residential to industrial. Secondly, to reflect local commuting, we include population density per the administrative unit and the local number of addresses, including industrial and commercial buildings, per the corresponding basic administrative unit. Thirdly, we have further enhanced this data to include information on vehicles commuting to the wider area of town to reflect long-haul commuting factors. In connection to the charging data, we have performed additional analysis of charging load data, uncovering insights that may be found in Appendix \ref{sec:app-b}.

\subsection{Model architecture}
There are two distinct, yet related components that DSOs are interested in, the first being peak power demand and the second being the actual load shape during the day \cite{nguyen2014charging, hashemi2019stochastic}. 
Moreover, our hypothesis is that, in general, there exist \textit{K} underlying archetypal charging profiles, which are mixed by factors depending on the spatiotemporal properties of a charging station. Once enough data flows into the model given by a neural network, our aim is to interpret and examine those profiles and how the mixture factors depend on particular spatiotemporal settings. 


\section{Results}

\begin{figure}[H]
    \centering
    \includegraphics[width=0.9\linewidth]{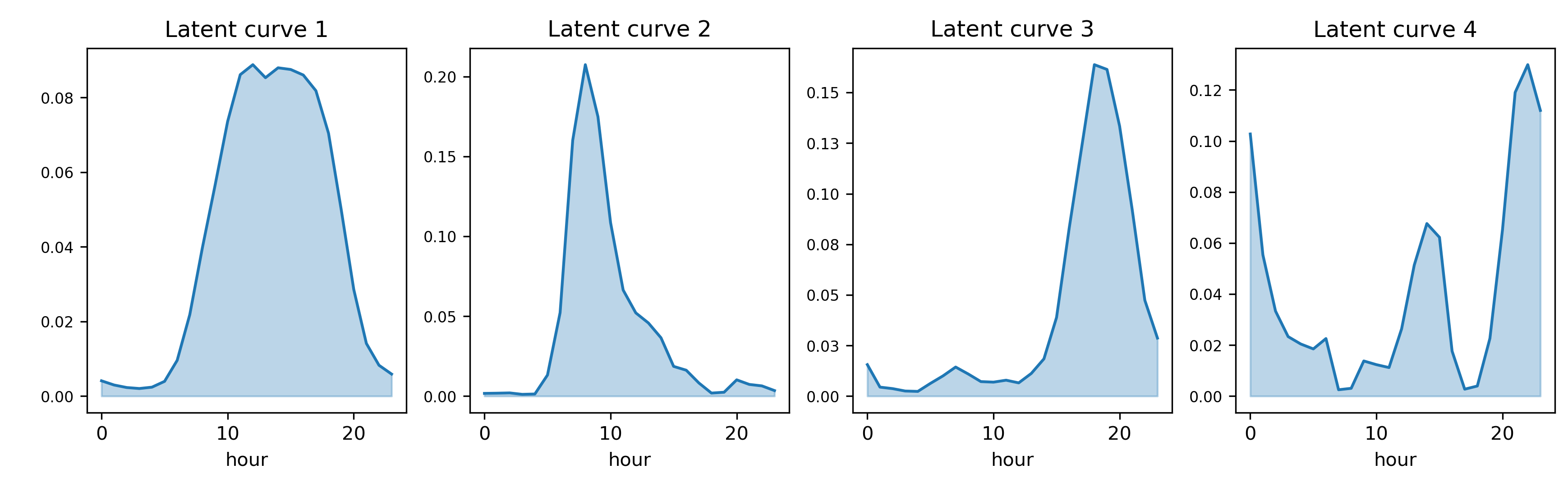}
    \caption{These 4 plots showcase trained latent load curve shapes within the neural network architecture. Note that the latent profiles are probability distributions, and thus the sum of the area they define is equal to 1.}
    \label{fig:loss-fig}
\end{figure}

Using the described experimental setup, we can examine the generated load profiles in Figure \ref{fig:loss-fig}. It was our hypothesis that the model would be able to decompose the input parameter influences and assign them to latent load curves representing archetypal behavior. After some experimenting, we have chosen $K = 4$ latent profiles in this initial study. Since the generated load curves differ from each other starkly, it appears the model works as intended. 

A basic interpretation of the generated latent load profiles is that latent curve 1 shows a gradual charging curve with a peak during the day, typical for public charging. The latent curve 2 shows a single morning peak demand, and the latent curve 3 shows a single evening demand, which is generally more typical for private charging. Lastly, latent curve 4 shows a residual demand with multiple peaks. What are the causes of these archetypal behaviors is a topic of further research, however, if we follow our hypothesis more in-depth, we may also relate the emergent load curves in the context of observed area charging specificity described in Appendix \ref{sec:app-b1}. 

When comparing to the generated load profiles observed here, we can see similar load profiles coming out of the original charging data matched to types of basic administrative units. If we were to make this connection speculatively, it creates an implication that the latent curve 1 might represent the local, residential charging archetypal behavior, latent curve 2 might represent recreational commute, latent curve 3 suburban and long-range commute, and latent curve 4 might correspond to irregular industrial and commercial commute. Based on this connection, it appears that out of the assembled input data, the strongest impact on the predicted load curves is held by the type of the local basic administrative unit.

\section{Conclusion and further work}

In this study, we have introduced a novel approach to creating a machine learning based model which allows for generative creation of public EV charging load profiles in various urban geographies given a wide range of input data. Our method is designed to create internal latent charging profiles, which have the potential to better explain the sources of charging demand if interpreted in connection with archetypal behaviors. In our results, we have offered such an interpretation based on available data analysis.

Our proposed model has the potential to assist DSOs and the wider research community in better modelling public charging demand, model different potential future scenarios, and in the case of DSOs, create more efficient strategies to expand EV charging grid infrastructure and support the EV number increase and its wider accessibility while keeping unexpected grid impacts to a minimum. Our model is still undergoing further refinement and there are some limitations to acknowledge. Firstly, we aim to include more data in our future iteration, as our current data has a cutoff in 2022 as described in Appendix \ref{sec:app-a2}. Connected to this, further investigation is needed to understand what effect the COVID pandemic lockdowns had on our input data, since as shown in Appendix \ref{sec:app-b5-covid}, the decrease in total load was significant. We are also continuing our research in the interpretability of the charging load profile sources, facilitated in our model via the latent archetypal load shapes, with the interpretation being speculative. An alternative approach might be to strictly separate types of inputs to be used for each latent load shape. Lastly, we invite the broader community to apply similar approaches to different geographies to cross-reference our findings.


\section*{Acknowledgements}
The authors would like to thank and acknowledge PREdistribuce, the Prague Distribution System Operator(DSO), for the charging data used in this study, and the Czech Statistical Office for data on demography and geography classification in Czechia. This work was supported by grants number SGS24/093/OHK5/2T/13 and SGS23/117/OHK5/2T/13 provided by CTU Prague, and number TS01020030 provided by the Technology Agency of the Czech Republic. Authors declare no conflict of interest.


\bibliography{arxiv}{}
\bibliographystyle{unsrt}

\appendix
\newpage
\section{Prague EV public charging landscape}
\label{sec:app-a}

\subsection{Prague area classification per basic administrative unit}
\label{sec:app-a1}
\begin{figure}[H]
    \begin{center}
        \includegraphics[width=0.7\linewidth]{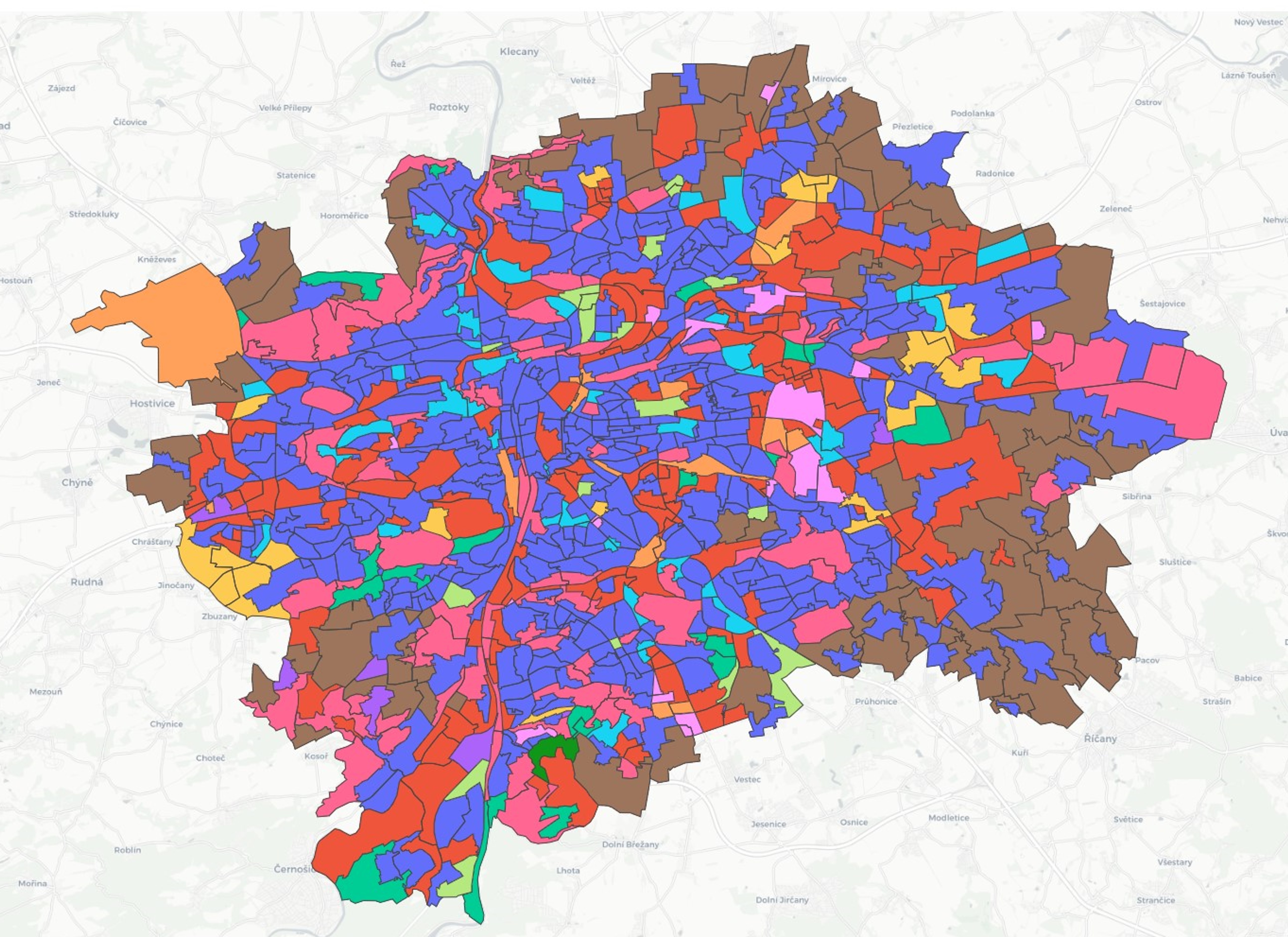}
    \end{center}
    \caption{ZSJ categories found in Prague proper}
    \vspace{-2em}
    \label{ZSJ-fig}
\end{figure}

In order to analyse the potential factors affecting charging demand, we have to understand the environment in which individual charging points are located. For this aim, we have utilized the enhanced "multi" layer of Basic Administrative Units (Základní sídelní jednotka, ZSJ), as provided by the Czech Statistical Office\cite{czech_statistical_office_zakladni_2024}. The ZSJ is the smallest administrative unit available, and there are 948 individual units in Prague. Using this data, we have been able to classify individual ZSJ areas in Prague into 12 categories, as seen in Figure \ref{ZSJ-fig}. 

Since the names for ZSJ categories are standardised in Czech, table \ref{ZSJ-translation} shows English translations and a colour-coded legend for the following figures. 
    \vspace{-0.5em}
\begin{table}[h]
    \begin{center}
    {
    \renewcommand{\arraystretch}{1.4}
    \tabcolsep=.1cm
    
    \begin{tabular}{|p{0.38\linewidth}|p{0.38\linewidth}|p{0.15\linewidth}|}
    \hline
    \multicolumn{3}{|c|}{Basic Administrative Unit (ZSJ) Categories} \\
    \hline
    Original Czech name &English translation    & Colour\\
    \hline
    
    \hline
    Obytná plocha v kompaktní zástavbě
        &Compact residential area
        &\cellcolor[HTML]{636efa}Purple\\
        
    \hline
    Městská a příměstská smíšená plocha
        &Urban and suburban mixed area 
        &\cellcolor[HTML]{EF553B}Red\\
        
    \hline
    Obytně rekreační plocha
        &Residential and recreational area 
        &\cellcolor[HTML]{00CC96}Teal\\
        
    \hline
    Odloučená obytná plocha
        &Separated residential area
        &\cellcolor[HTML]{AB63FA}Violet\\
        
    \hline
    Dopravní areál
        &Transportation infrastructure area  
        &\cellcolor[HTML]{FFA15A}Orange\\

    \hline
    Areál občanské vybavenosti
        &Civic amenities area 
        &\cellcolor[HTML]{19D3F3}Blue\\

    \hline
    Rekreační plocha
        &Recreational area  
        &\cellcolor[HTML]{FF6692}Coral\\

    \hline
    Ostatní účelová plocha
        &Urban and suburban mixed area  
        &\cellcolor[HTML]{B6E880}Lime\\

    \hline
    Průmyslový areál
        &Industrial area  
        &\cellcolor[HTML]{FF97FF}Pink\\

    \hline
    Rezervní plocha
        &Reserve area  
        &\cellcolor[HTML]{FECB52}Yellow\\

    \hline
    Zemědělská plocha
        &Agricultural area  
        &\cellcolor[HTML]{9E7658}Brown\\

    \hline
    Lesní plocha
        &Forest area  
        &\cellcolor[HTML]{0D9816}Green\\

    \hline
    \end{tabular}}
            \vspace{1em}
    \caption{Legend for ZSJ categories in following figures.}
            \vspace{-4em}
    \label{ZSJ-translation}
    \end{center}
\end{table}
     \vspace{-1em}

\newpage
     
\subsection{Charging demand localization in Prague}
\label{sec:app-a2}
\begin{figure}[H]
    \begin{center}
        \includegraphics[width=0.8\linewidth]{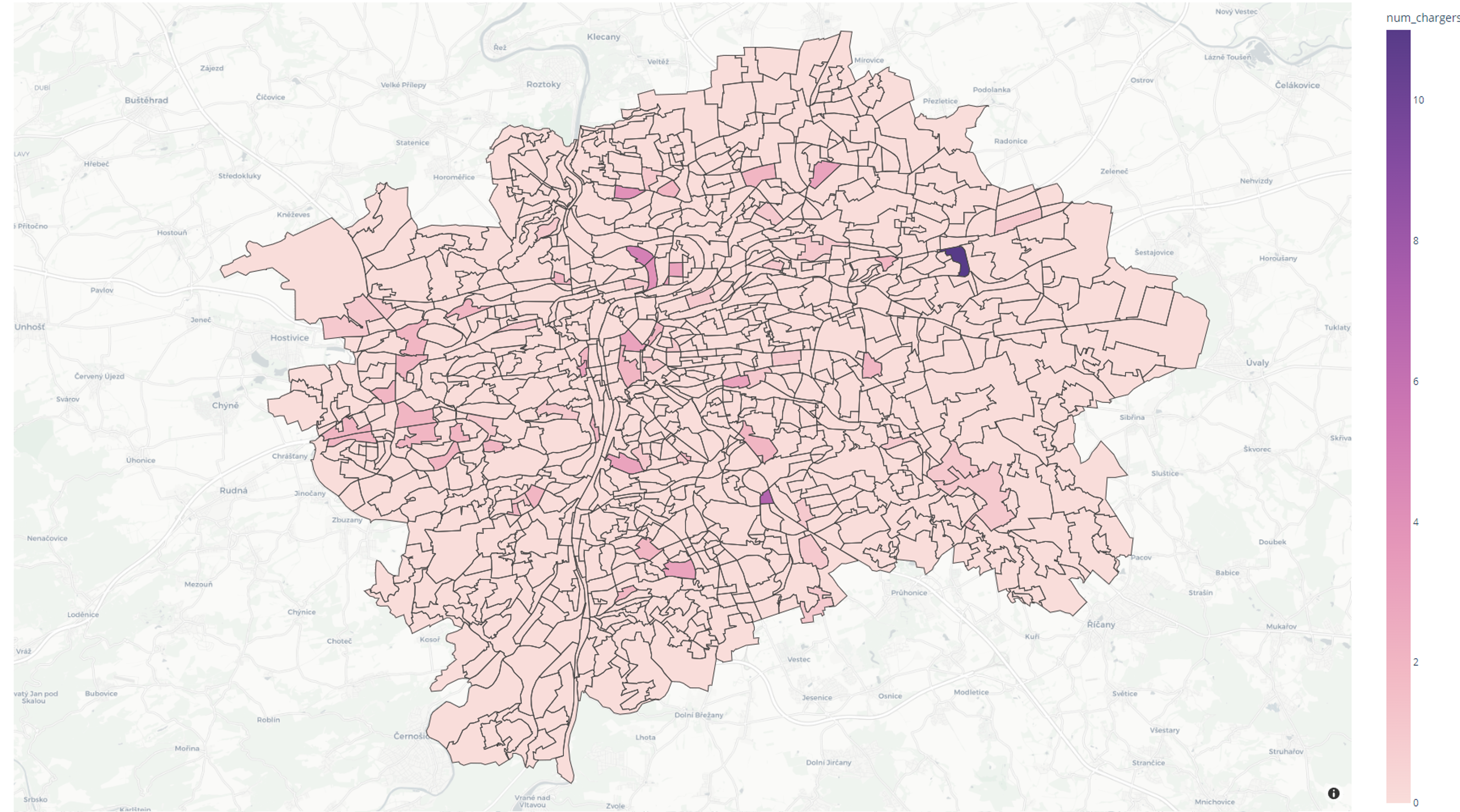}
    \end{center}
                    \vspace{-0.5em}
    \caption{Heatmap of public charging point locations per ZSJ in Prague based on the available data.}
                \vspace{-1em}
    \label{charger-fig}
\end{figure}

The data on individual charging sessions was kindly provided by Prague's largest provider of public charging points, PREdistribuce\cite{predistribuce_verejne_2024}. Figure \ref{charger-fig} shows a visualisation of charger locations within the ZSJ structure of Prague.

\begin{figure}[H]
    \begin{center}
        \includegraphics[width=0.9\linewidth]{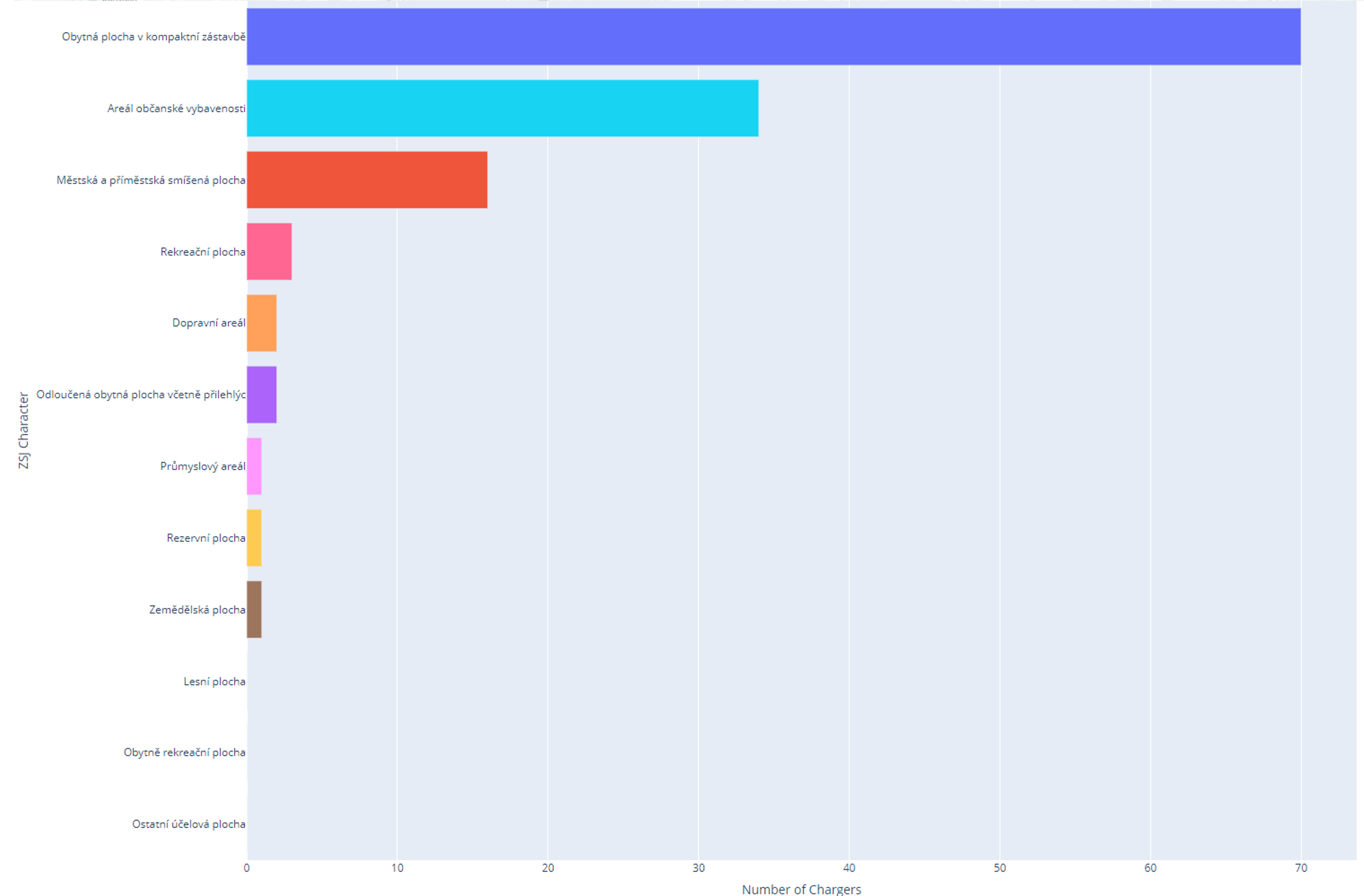}
    \end{center}
    \caption{Number of chargers per ZSJ category across Prague}
        \vspace{-2em}
    \label{ZSJ-charge}
\end{figure}

We can see the broad distribution of public chargers across the city of Prague, with the largest concentration in population centres, most notably the one at Černý most in the top right of the map, with 12 chargers present in one ZSJ. In Figure \ref{ZSJ-charge}, we can see that most chargers in the dataset are placed in residential areas (70 chargers), followed by civic amenities areas (34 chargers) and urban and suburban mixed areas (16 chargers). The other categories of ZSJ areas have 3 or fewer chargers present.
\vspace{-1em}

\begin{figure}[H]
    \begin{center}
        \includegraphics[width=1\linewidth]{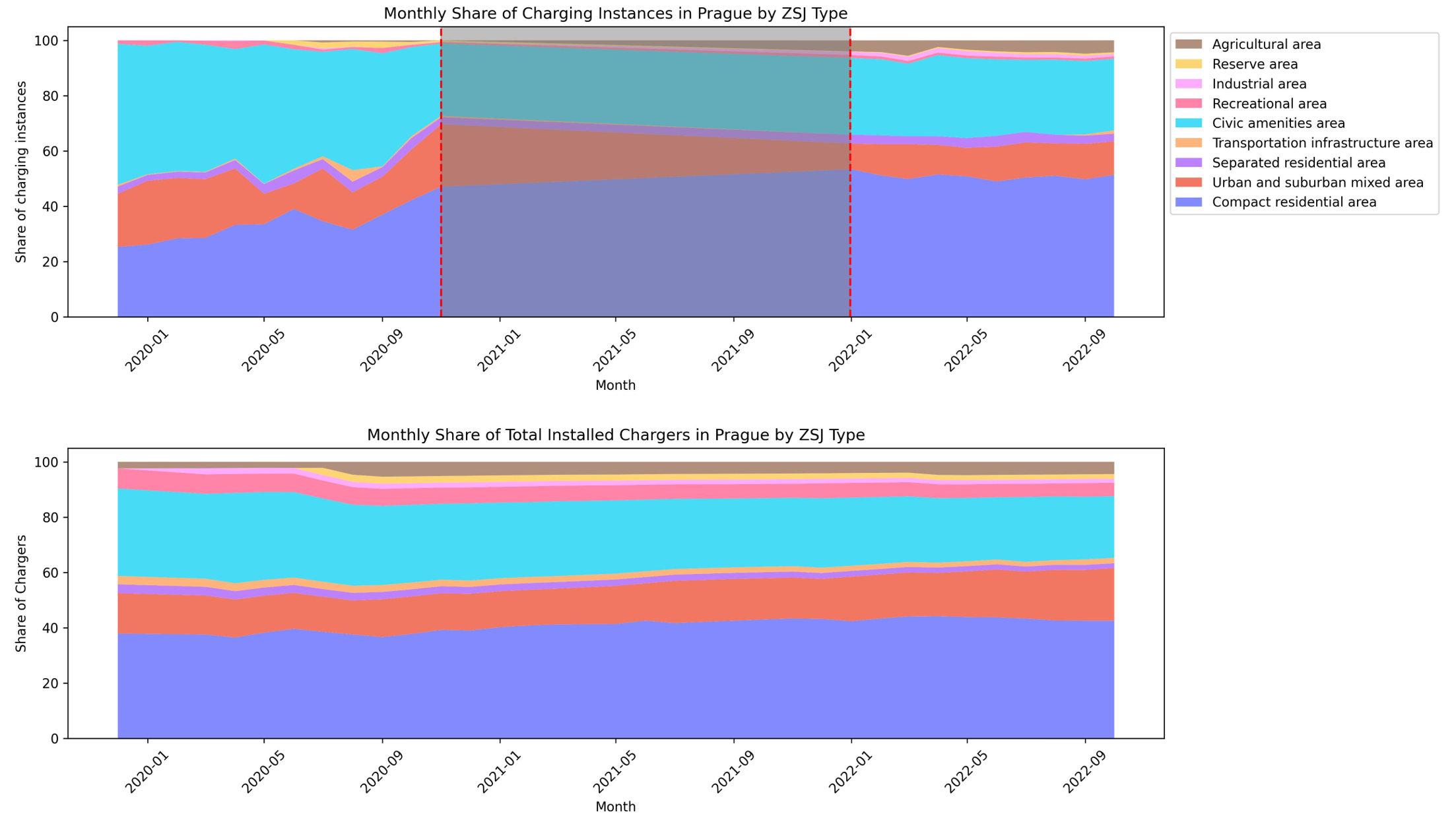}
    \end{center}
    \caption{Temporal relative share development analysis for charging instances and installed chargers in Prague, classified per ZSJ type. Note that the red-bordered gray fill area in the upper chart of charging instances corresponds to the interpolated region of unavailable data, and the lower chart of the number of installed chargers includes chargers put into operation before the span of the chart timeline, December 2019. 
}
        \vspace{-2em}
    \label{fig:load-development-ZSJ}
\end{figure}

In Figure \ref{fig:load-development-ZSJ}, the monthly relative total shares of each ZSJ area type are visualized for two key elements in our dataset: charging instances and installed charging points. Note that for the charging instances dataset, no datapoints are currently available to us between December 2020 and December 2021, as highlighted in the figure by a red-bordered gray fill-in with interpolated data. We can see that the number of charging sessions in Compact residential areas rose between the years 2019 and 2022. Simultaneously, a modest increase in the relative share of charging instances can be observed in Agricultural and Industrial areas. Conversely, a decline in charging instances share is evident in Civic amenities, Urban and suburban mixed, and Recreational areas. This decrease could be attributed to various factors, such as changes in usage patterns or the availability of charging stations.

However, combined with the analysis of the share of installed charging stations on the lower plot, we can see that the change in charging sessions relative shares for individual ZSJ area types described above are not directly explainable by developments in the number of available chargers in the respective ZSJ types. Indeed, we can see that while the two most dominant ZSJ types (residential areas and civic amenities areas, respectively) are overrepresented in the number of charging instances, the suburban, recreational, transportation, and reserve areas are greatly underrepresented in the charging instances relative to the number of available chargers. This observed behavior may be caused by several characteristics of chargers in the dataset, such as accessibility, geographic location, charging point alternatives, and others. Still, this discrepancy clearly implies a different behavioral pattern for chargers in different ZSJ area types, which is examined next.

\newpage

\section{Public charging load curve analysis}
\label{sec:app-b}

\subsection{Load-curve analysis based on basic administrative unit type}
\label{sec:app-b1}

\begin{figure}[H]
    \centering
    \includegraphics[width=0.4\linewidth]{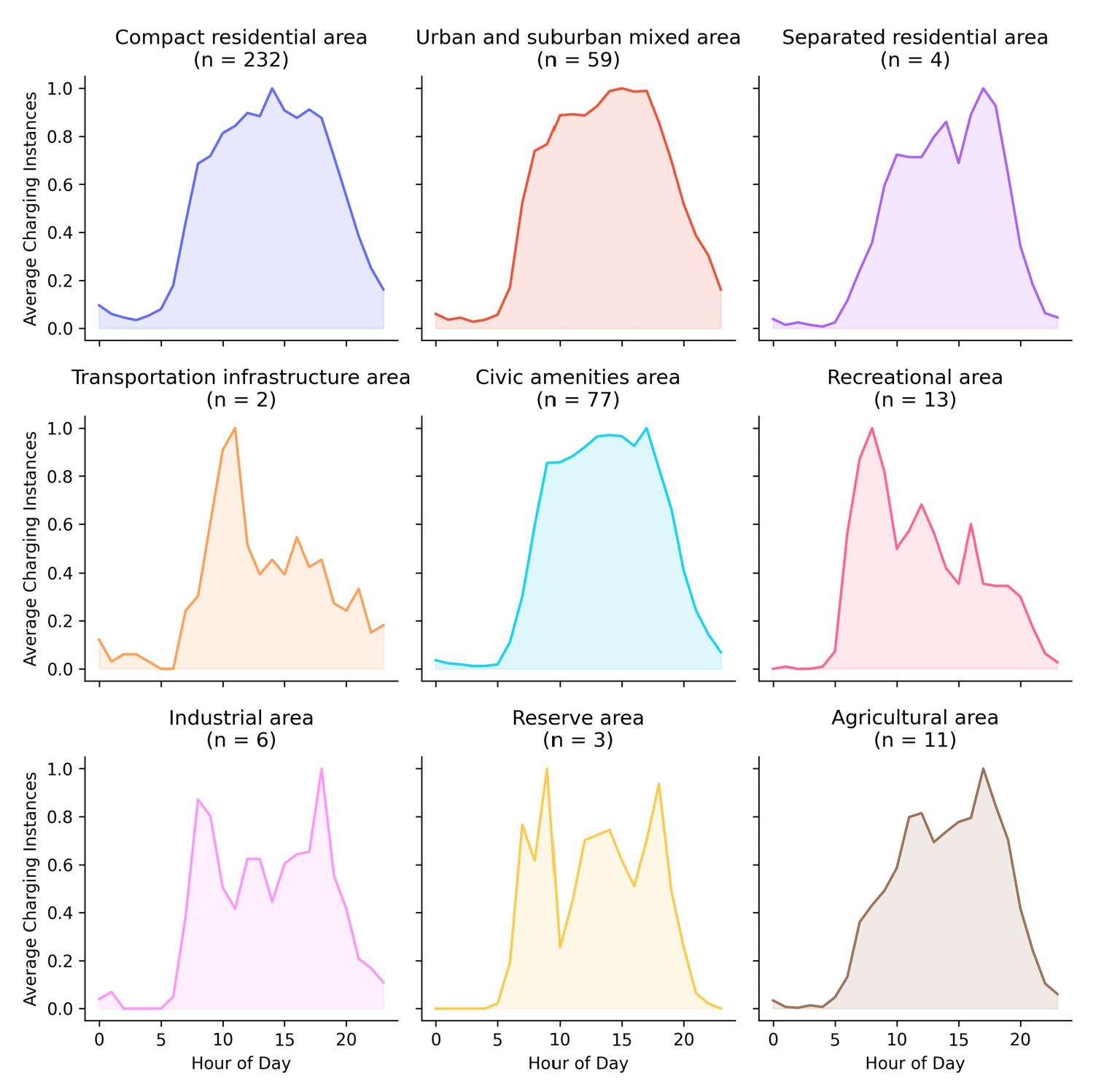}
    \caption{Normalized demand curves for ZSJ types with \textit{n} defined as number of charge points present in the dataset for each ZSJ type}
    \label{fig:load-zsj-fig}
\end{figure}

We have investigated the specificity of ZSJ area types by plotting demand curves of EV charging split per individual type. Figure \ref{fig:load-zsj-fig} provides a visualization of the normalized average number of charging instances per hour of day, which approximates load curve behavior patterns across different ZSJ area types. The analysis clearly shows that demand patterns between different ZSJ types are heterogeneous. We see four distinct groups emerge:

\begin{itemize}[nosep]
    \item \textbf{Group 1: Sustained single peak areas} demonstrate a sustained demand in charging instances with a gradual single peak throughout the day, suggesting a consistent demand for charging facilities in these areas. The following ZSJ types fall within this category:
    \begin{itemize}[nosep]
        \item Compact residential area
        \item Urban and suburban mixed area
        \item Civic amenities area
    \end{itemize}
    \item \textbf{Group 2: Morning single peak areas} exhibit a clear peak in the morning hours, around 8:00. The following ZSJ types fall within this category:
    \begin{itemize}[nosep]
        \item Transportation infrastructure area
        \item Recreational area
    \end{itemize}
    \item \textbf{Group 3: Evening single peak areas} demonstrate a single peak in the evening, around 17:00. The following ZSJ types fall within this category:
    \begin{itemize}[nosep]
        \item Separated residential area
        \item Agricultural area
    \end{itemize}
    \item \textbf{Group 4: Double peak areas} demonstrate both the morning peak and the evening peak, though potentially shifted by an hour. The following ZSJ types fall within this category:
    \begin{itemize}[nosep]
        \item Reserve area
        \item Industrial area
    \end{itemize}
\end{itemize}

In the figure we compare data within time series or categories of inherently different volume, in categories of locations with varying numbers of charging points, we have therefore utilized max normalization feature scaling. It rescales all values of a non-negative data series to a maximal value of one, while retaining the relative distance from zero. Normalized series and their characteristics are then directly comparable. The \(i\)-th element of the series is rescaled as follows:

\[
x_i' = \frac{x_i}{\max(x)}
\]

\newpage
\subsection{Average load curves per month and weekday}
\begin{figure}[H]
    \centering
    \includegraphics[width=0.8\linewidth]{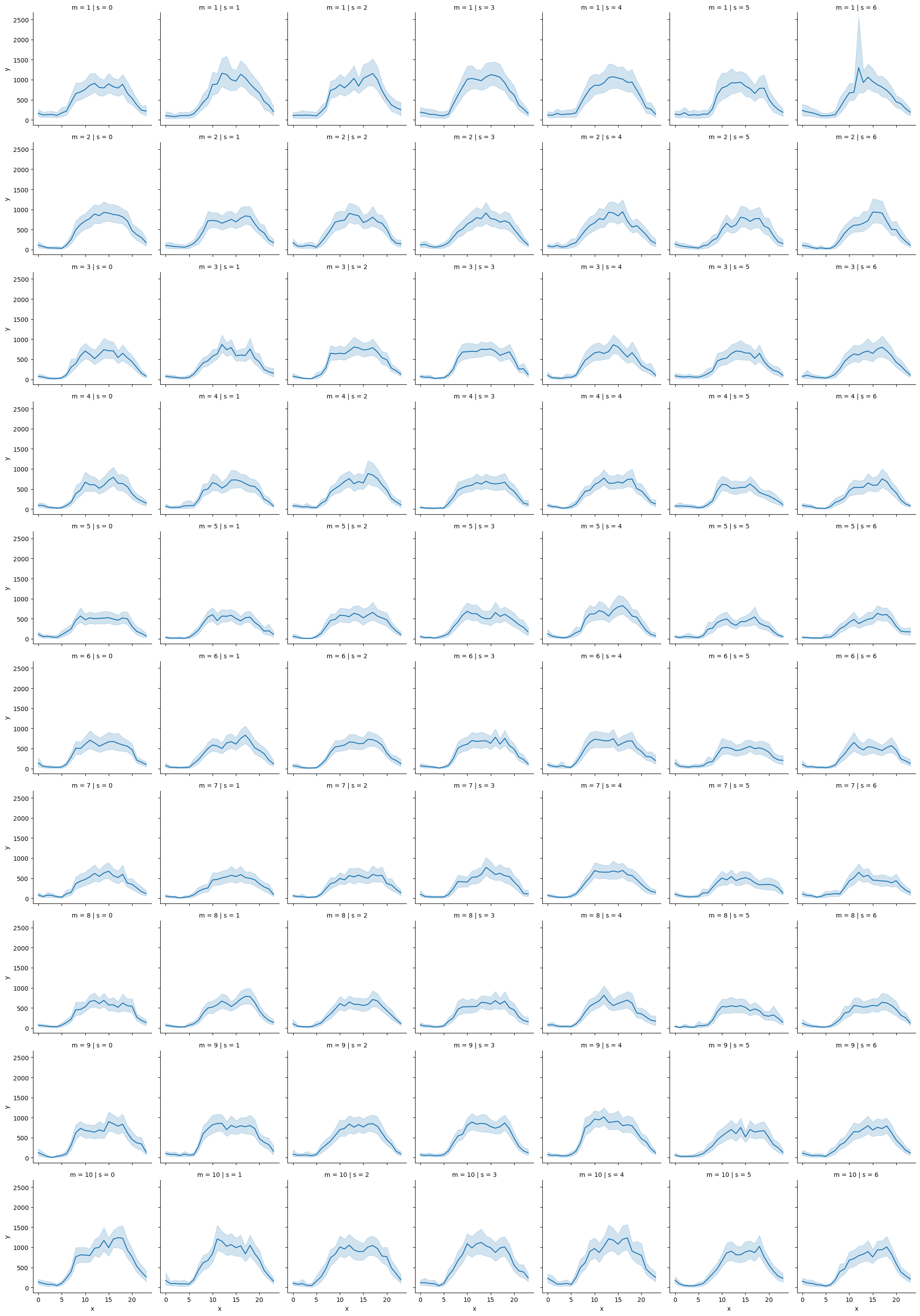}
    \caption{Comparing the average load curves and their confidence intervals for each weekday (columns Monday to Friday) and month (rows January to December)}
    \label{fig:monthweek-fig}
\end{figure}

We have also observed trends in seasonality between weekdays and months as shown in Fig. \ref{fig:monthweek-fig}. While no significant load shape differences are found, we can observe marked lower demand on the weekends and in summer months.

\label{month-workweek}

\newpage
\subsection{Weekday analysis}
\begin{figure}[H]
    \centering
    \includegraphics[trim=0cm 0cm 0cm 90.6cm, clip, width=0.5\linewidth]{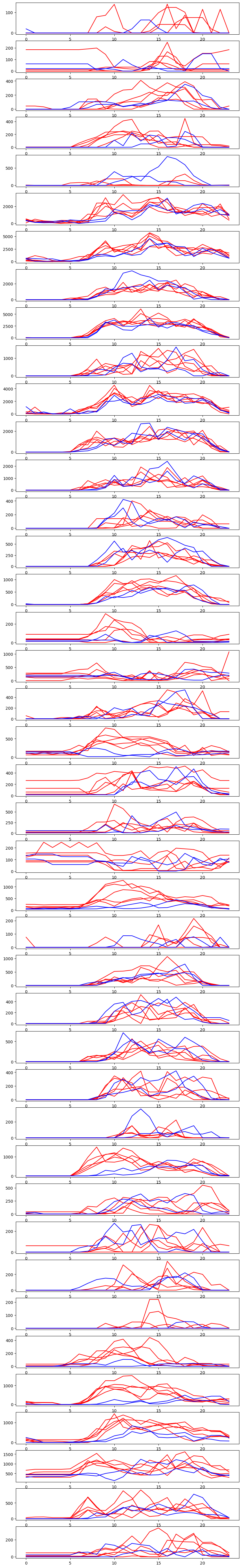}
    \caption{Comparing load curves for weekends (marked in blue) and weekdays (marked with red) for a subsample of individual chargers}
    \label{fig:weekday-fig}
\end{figure}

Further delving into the differences between individual weekdays, in Fig. \ref{fig:weekday-fig} we have shown individual load-curve shapes between weekdays and weekends for a subset of chargers in the dataset. While weekends show generally lower demand, high variability is also observed.
\label{weekday}

\newpage
\subsection{Easter holiday analysis}
\begin{figure}[H]
    \centering
\includegraphics[width=0.99\linewidth]{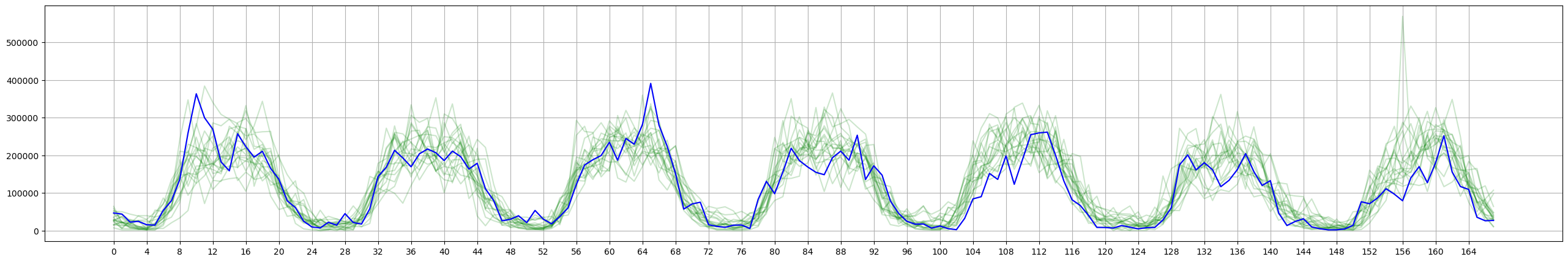}
    \caption{Easter week average charging load curve in blue compared to average loads of standard workweeks in green}
    \label{fig:easter-fig}
\end{figure}

Following our analysis in \ref{month-workweek} and  \ref{weekday}, we have examined what effect public holidays might have on charging, by comparing the Easter holiday week average load curve with other weeks shown in Fig. \ref{fig:easter-fig}, however no major differences or anomalies were found.

\subsection{COVID lockdown total load impact analysis}
\label{sec:app-b5-covid}
\begin{figure}[H]
    \centering
\includegraphics[width=0.99\linewidth]{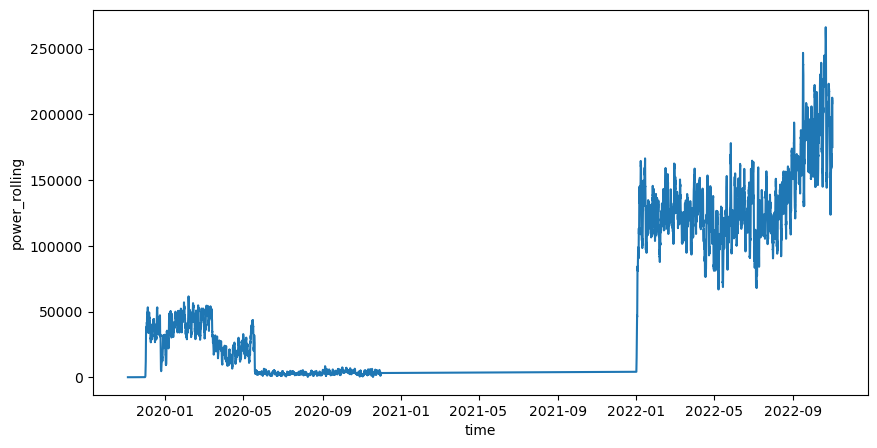}
    \caption{Total load observed during the dataset}
    \label{fig:covid-fig}
\end{figure}

As shown in Fig. \ref{fig:covid-fig}, COVID lockdowns in Prague after May 2020 have had a stark impact on general public charging loads. The impact on our dataset is a topic of further research.
\newpage
\section{Hyperparameters used for running the Experiment}
\label{sec:app-c}

The following table lists the hyperparameters used to configure and train the \texttt{ChargingProfileModel} in the experiment described in this paper:

\begin{table}[h!]
\centering
\renewcommand{\arraystretch}{1.5} %
\begin{tabular}{|l|c|}
\hline
\textbf{Hyperparameter} & \textbf{Value} \\
\hline
Charging Profile Granularity & 24 \\
Hidden Size & 128 \\
Hidden Size (g) & 64 \\
Latent Profiles Count & 4 \\
Loss Function & \texttt{nn.MSELoss(reduction="mean")} \\
Optimizer & \texttt{torch.optim.Adam} \\
Learning Rate & 0.0004 \\
Number of Epochs & 150 \\
\hline
\end{tabular}
\vspace{0.5cm} 
\caption{Hyperparameters used in the ChargingProfileModel experiment.}
\label{table:hyperparameters}
\end{table}

In our experiment, we used the \texttt{ChargingProfileModel} with the following hyperparameters:

\begin{itemize}
    \item \textbf{Charging Profile Granularity}: Set to 24, this parameter defines the granularity of the charging profile, corresponding to the daily granularity of a 24 hour cycle in our case.
    \item \textbf{Hidden Size}: The size of the hidden layers in the \emph{f module}, set to 128.
    \item \textbf{Hidden Size (g)}: The size of the hidden layers in the \emph{g module}, set to 64.
    \item \textbf{Latent Profiles Count}: The number of latent profiles, set to 4 in our experiment run.
    \item \textbf{Loss Function}: Mean Squared Error (MSE) loss function, specified as \texttt{nn.MSELoss(reduction="mean")}.
    \item \textbf{Optimizer}: Adam optimizer, used for updating the model parameters.
    \item \textbf{Learning Rate}: Set to 0.0004, this controls the step size during optimization.
    \item \textbf{Number of Epochs}: The model is trained for 150 epochs.
\end{itemize}

These hyperparameters were chosen to optimize the performance of the \texttt{ChargingProfileModel} on the given dataset. The model was trained using the Adam optimizer with a learning rate of 0.0004 and the MSE loss function. The training process was carried out for 150 epochs to ensure adequate learning.

\end{document}